# LLM experiments with simulation: Large Language Model Multi-Agent System for Simulation Model Parametrization in Digital Twins

Yuchen Xia, Daniel Dittler, Nasser Jazdi, Haonan Chen, Michael Weyrich
*Institute of Industrial Automation and Software Engineering*
*University of Stuttgart*
Stuttgart Germany
{yuchen.xia; daniel.dittler; nasser.jazdi; michael.weyrich}@ias.uni-stuttgart.de

*Abstract*— This paper presents a novel design of a multi-agent system framework that applies large language models (LLMs) to automate the parametrization of simulation models in digital twins. This framework features specialized LLM agents tasked with observing, reasoning, decision-making, and summarizing, enabling them to dynamically interact with digital twin simulations to explore parametrization possibilities and determine feasible parameter settings to achieve an objective. The proposed approach enhances the usability of simulation model by infusing it with knowledge heuristics from LLM and enables autonomous search for feasible parametrization to solve a user task. Furthermore, the system has the potential to increase user-friendliness and reduce the cognitive load on human users by assisting in complex decision-making processes. The effectiveness and functionality of the system are demonstrated through a case study, and the visualized demos and codes are available at a GitHub Repository: https://github.com/YuchenXia/LLMDrivenSimulation

## I. INTRODUCTION

Digital twin technology has significantly improved productivity in industries such as engineering and manufacturing by providing virtual replicas of physical systems. Digital twin system offers an infrastructure that enables **users** to monitor, simulate, predict, and control real-world processes in realtime, thereby improving decision-making and enhancing operational efficiency.[1]

Currently, users are responsible for interpreting data from digital twins, understanding complex system behaviors, and making informed decisions. Human users can only process a limited amount of information at a time, constraining the efficiency and usability of digital twin systems. Processing information requires cognitive effort and specialized knowledge, often necessitating extensive training to ensure accurate understanding and effective control of the processes that is represented in digital twins. Additionally, the availability of such skilled users can be limited, and the high knowledge barrier can lead to delays and compromised decision-making.

Large Language Models (LLMs) have demonstrated significant intelligence in understanding knowledge. Incorporating the intelligence of LLMs with digital twin technology allows for a higher degree of automation and system intelligence, presenting new opportunities to enhance digital twins.

Given the critical role users play in interacting with digital twins to monitor and predictively control processes, a pertinent question arises: Can this user role be assisted by large language models to automate the interaction with digital twin simulation, thus determining feasible parametrizations for controlling the simulated processes?

To answer this question, the key contributions of this paper include:

- **Multi-agent framework:** We present a novel LLM-agent architecture that dynamically interacts with digital twin simulation to determine feasible parameter settings for simulated physical processes.

- **Proof-of-concept:** A case study demonstrates the system's capability to automate simulation parametrization, employing heuristic knowledge reasoning from LLMs.

- **Applications potential:** the results suggest that integrating LLMs with digital twin technologies can enhance the intelligence and user-friendliness of industrial digital twin systems, improving operational efficiency and accessibility.

## II. BACKGROUND

Figure 1 selectively illustrates the key conceptual components in this work. These components include a physical entity under study, its virtual counterpart in the form of a digital twin, and an LLM multi-agent system designed for intelligent interaction.

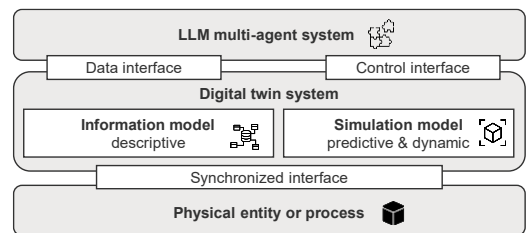

Figure 1 Key conceptual components in this work

### A. Digital twin

While definitions of digital twins differ in similar details [1][2], they converge on key aspects: A digital twin is a software system that represents a physical entity or system that mirrors real-world conditions, processes, and systems. It remains synchronized with its physical counterpart through interfaces that manage data acquisition and control physical processes (e.g., through an automation system) [3][4]. In the context of this paper, two primary parts of the digital twin are selectively considered:

**Information Model:** This component focuses on the structured integration and management of descriptive data. It serves as the core of data storage and retrieval within the digital twin system, facilitating access and processing of detailed information about the physical entity [5].

**Simulation model:** In contrast to the information model, the simulation component is predictive and dynamic. It is primarily used for operational testing and scenario analysis, enabling the exploration of potential outcomes based on various inputs and conditions.[6]

The digital twin serves as a critical digital interface for the LLM system. It provides LLM agents with access to information about physical processes, thereby enabling the LLM system to develop intelligent applications based on these data.

### B. "Divide and conquer" and LLM multi-agent system

Due to the complexity of the task or the complex data to be processed from the digital twin, a single LLM often cannot produce satisfactory results. To address this, complex tasks are decomposed into smaller, more manageable components. This "divide and conquer" approach allows multiple LLM agents to collaboratively tackle these tasks, leading to a comprehensive and effective solution. This principle can be realized by designing LLM multi-agent systems.

### C. Related works

Several studies have explored the development of LLM-powered applications, often applying the "divide and conquer" principle to design these systems. These can be mainly organized on two levels:

**Model level:** in related works, several frameworks have been developed to manage the complexity. The Chain-of-Thought (CoT)[7] framework facilitates systematic, step-by-step reasoning, while the Tree-of-Thought (ToT)[8] allows LLMs to explore multiple pathways, identifying the most viable solutions. Additionally, the ReAct framework [9] generates reasoning traces and task-specific actions in an interleaved manner, effectively addressing complex tasks. These frameworks can be realized by systematic prompting to guide the reasoning processes of LLMs, enhancing their capability to solve intricate problems.

Alternative methods involve fine-tuning the LLM to modify its text generation behavior. Techniques such as Proximal Policy Optimization (PPO)[10] and Direct Policy Optimization (DPO)[11] adjust the generation policies of the LLM, enhancing its ability to produce logically connected texts and thereby improving its problem-solving capabilities.

**Agents level:** Recent studies [12],[13] have explored frameworks to design LLM agents playing different roles in task-oriented coordination within simulated social environments, emphasizing reflection and decision-making processes. Another study introduces a framework that enables LLMs to act as game players controlling real-time strategy games by analyzing complex data and managing game dynamics [14]. In our prior work, we developed a multi-agent system to autonomously plan and control automation systems via a digital twin system [3]. Additionally, a comprehensive survey summarized in [15] reviews recent advancements in LLM multi-agent systems.

In the following section, we present a novel multi-agent system framework to interact with simulation models in digital twin system to perform simulated experiments to determine satisfactory parametrization of a process.

## III. THEORETICAL FRAMEWORK OF A MULTI-AGENT SYSTEM

In this work, we propose a structured framework for multi-agent system with that assigns distinct roles and responsibilities to each LLM agent within the information processing pipeline. This framework includes four types of agents: observation, reasoning, decision and summarization.

**Observation agent:** This agent collects and observes real-time data from the digital twin, focusing on operational parameters, current conditions, and changes in key metrics crucial for optimal outcomes. It identifies pertinent observation data, filters out noise and irrelevant information, and distills important insights to establish a factual foundation for further analysis.

**Reasoning agent:** Incorporating data preprocessed by the observation agent, the reasoning agent interprets and analyzes the observed data and insights. It generates reasoning steps toward actionable decisions. This process can be characterized as heuristic reasoning, as it utilizes knowledge patterns that the LLM has acquired from extensive training data.

**Decision agent:** Upon completion of the reasoning phase, the decision agent generates executable actions based on the previously generated intermediate results. These generated actions can be formulated as API calls, procedure calls or services in a software system to invoke operations.

**Summarization agent:** The summarization agent consolidates the outcomes of the observation, reasoning, and decision processes. It compiles a concise report highlighting the significant insights and results, providing an overview of system performance. This agent helps users quickly understand key information about the system's behavior.

This multi-agent system design framework resembles the scientific methodology used in empirical experiments. It provides a structured and reproducible approach for creating LLM multi-agent systems. By clearly defining roles and responsibilities in information processing, this framework aids in the systematic development, rigorous testing, and iterative improvement of systems powered by LLMs.

## IV. METHODOLOGY

### A. System overview

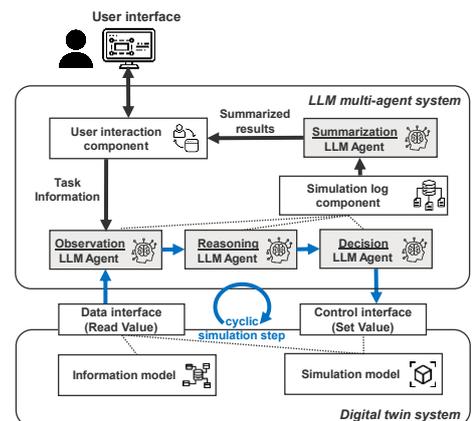

Figure 2 Architecture overview of LLM multi-agent system interacting with simulation model in a digital twin

As shown in Figure *2*, this architecture comprises three layers. At the top layer, the user interface receives user requirements and the objectives of the simulation, which are set as target goals for the simulation to achieve. In the middle layer, an LLM multi-agent system interacts with digital twins via data and control interfaces. At the fundamental layer, the digital twin is structured to update and operate step-by-step in a cyclic manner. During each cyclic simulation step, data is accessed through a data interface, and simulation parameters are adjusted via a control interface.

The **observation agent** processes the information from the data interface to extract significant insights and distill crucial aspects from the extensive data retrieved, setting the stage for the next agent to achieve targeted goals. Subsequently, the **reasoning agent** applies heuristic reasoning to analyze the situation and deduce the next viable control strategies, and the **decision agent** generates actions in the form of parametrized function calls to adjust the simulation's parameters for the upcoming step based on the previously generated observation and reason in text. Collectively, the agents monitor the cyclic steps of the simulation and control its progression.

Data from each cycle and the corresponding control parameters are recorded in a simulation log component, which captures detailed information about the LLM agents' iterative decision-making processes. The **summarization agent** compiles these logs for each simulation step to develop a concise parametrized sequential control plan, which will be reported to the user as a feasible solution to a user task.

### B. Information representation conversion between digital twins and LLM agents

#### 1) Information processing of LLM agents

In this multi-agent system, each agent functions as an information processing component. A specific prompt is crafted for each agent to direct its behavior according to the sub-task it handles within the multi-agent framework (cf. Figure *3*). Both input and output information are converted into textual form. This conversion requires a translation process that transforms modeled information from the digital twin simulation model into a textual knowledge representation. This representation is then integrated into a prompt template. We apply this method to program the LLM multi-agent system in the case study section.

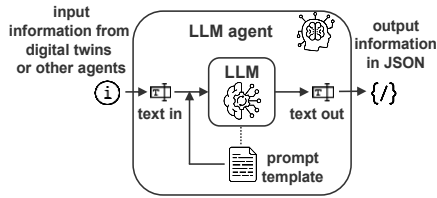

Figure 3 LLM agent as an information processor through Prompting

#### 2) JSON and function call as agent output

The generated text is converted into JSON format to facilitate technical processing by the software. At the end of the agent information processing pipeline, the decision agent produces a parameterized function call that serves as an actionable decision. This function call is managed by the simulation model's interface, which executes the operations within the simulation model for the next simulation step.

## V. A CASE STUDY AND IMPLEMENTATION

For a practical demonstration of the concept, we implemented a simplified container mixing process. This simulation is akin to mixing process in a blender, where the objective is to achieve a homogeneous mixture of balls (or particles) with varying densities. The balls are added to the container in a controlled sequence, and then the container is shaken to redistribute the contents.

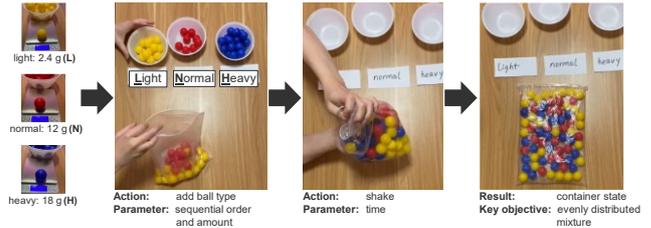

Figure 4 The simulated physical process

The case study experiments, depicted in Figure *4*, demonstrate how the order of adding the balls (light, normal, heavy) and the duration of shaking affect the mixture's homogeneity.

| Experiments | #1 | #2 | #3 | #4 |
|---|---|---|---|---|
| Parameter setting | Order: L – N – H  Shake: 7 seconds | Order: H - L - N  Shake: 10 seconds | Order: 1/2 L - 1/2 N - 1/2 L - 1/2 N - H  Shake: 5 seconds | Order: L - N - H  Shake: 40 seconds |
| Result |  |  |  |  |
| Objective: To achieve homogenous mixure | good |  | good |  |

Figure 5 The experimenting results with the physical process

We parametrized these variables to explore different combinations and their impact on the mixing outcome. Several experiments' parameter settings and results are summarized, as shown in Figure *5*.

### A. Simulation model for the case study

In the simulation, we model a container as a 10x10 matrix where two primary actions are possible: adding different types of balls and shaking the container. With each shake of the container, there is a probability of heavier balls switching positions with nearby lighter balls beneath. The simulation begins with the user setting a goal to achieve "an evenly distributed mixture", as shown in Figure *6*. The interaction with the LLM multi-agent system is facilitated through a chat box, and the simulation's progress is visualized on the user interface.

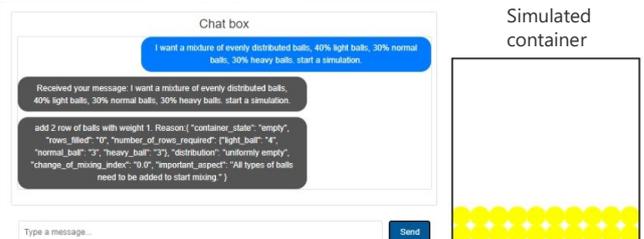

Figure 6 The user interface with chat box and visualized simulation

The LLM multi-agent system processes the user command, observing, reasoning, and controlling the simulation steps to achieve the desired outcome. The final result, including a visualization of the outcome and a summary of the control sequence, is then presented to the user in a concise format, as illustrated in Figure *7* and the demo.

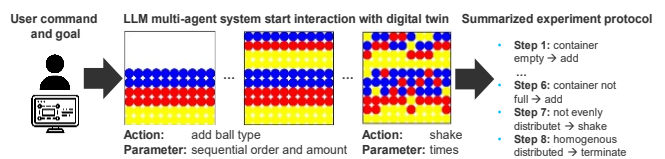

Figure 7 The LLM enhanced simulation execution process

## B. Knowledge representation in text

To realize this application, a key aspect is the conversion of knowledge representation: translating the simulation representation into a textual format that the LLM can interpret while preserving the knowledge. In the case study, the state of the container is represented as a matrix, where each position within the matrix is denoted by a number, as shown in Figure 8. Each number corresponds to a specific type of ball, which is explicitly defined within the agent's prompt. This textual conversion is critical for allowing the LLM to perceive and interact with the simulation data meaningfully.

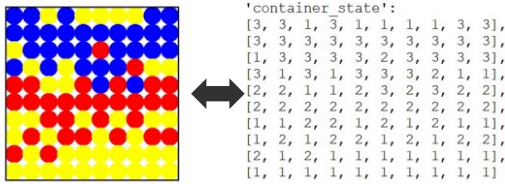

Figure 8 Conversion of simulation information into textual knowledge representation

To enhance the LLM's ability to assess the distribution within the mixture, we have integrated a quantifiable proxy metric. This metric quantitatively measures the degree of homogeneity in the distribution. After calculation, this indicator is also integrated into the prompt:

$$Degree\ of\ even\ distribution = \frac{\sum Diversity\ of\ Neighbor}{Number\ of\ balls}$$

## VI. RESULTS AND DISCUSSION

We developed this system, the code and demo video is released on the GitHub repository. The LLM operates similarly to a human experimenter within a virtual environment, interfacing with a digital twin to achieve an objective.

| Simulation run | #1 | #2 | #3 | #4 |
|---|---|---|---|---|
| Parameter setting | Order: 40% L – 20% H – 30% N – 10% H Shake: 5 times | Order: 20% L – 20% N – 20% H – 20% L – 10% N – 10% H Shake: 1 time | Order: 20% L – 20% N – 20% H – 20% L – 10%N Shake: 2 times | Order: 40% L – 20% H – 30% N – 10% H Shake: 5 times |
| Result | | | | |

Figure 9 The outcomes of experiments from the interaction between LLM and digital twin simulation

The system is capable of executing multiple simulation runs, allowing the LLM multi-agent system to explore a range of parameter settings and determine the most effective configurations, as demonstrated in Figure 9. This iterative process mirrors the method by which humans apply heuristic reasoning to experiment with various configurations in search of an optimal solution.

The integration of LLMs into the digital twin framework offers several advantages. Firstly, it enhances user experience by making interactions more intuitive and automating the parametrization process. Secondly, it lowers the knowledge barrier, making advanced digital twin technologies accessible to a wider range of users. This automation significantly reduces the need for manual effort in interpreting complex details and performing reasoning tasks, thereby accelerating the entire process. Moreover, utilizing a simulated environment for testing provides a risk-free scenario analysis, which minimizes operational risks associated with real-world testing. Finally, the combined predictive capabilities of LLMs and the simulation framework allow for more efficient strategic planning and process control.

## VII. CONCLUSION AND OUTLOOK

This research demonstrates the viability of incorporating Large Language Models into digital twin frameworks to automate the parametrization of simulations. By developing an LLM multi-agent system that interacts with the data and control interfaces of the digital twin, we can create more intelligent and user-friendly applications. The iterative, heuristic-based approach utilized by LLM agents mirrors the human problem-solving process and aims to generate real-time operational strategies for process control within a simulation environment. Moving forward, we plan to conduct further systematic testing and evaluations, refine the system, and investigate the LLM multi-agent framework on more sophisticated simulation models. We expect that continued advancements will broaden the application scope of LLM-enhanced digital twins, potentially leading to more efficient and productive systems across various industrial sectors.

### ACKNOWLEDGMENT

This work was supported by *Stiftung der Deutschen Wirtschaft (SDW)* and the Ministry of Science, Research and the Arts of the State of Baden-Wuerttemberg within the support of the projects of the *Exzellenzinitiative II*.

Special thanks to Yuye Tong for assisting with the preparation of the experiment materials.

### REFERENCES

[1] D. Dittler, D. Braun, T. Müller, V. Stegmaier, N. Jazdi, and M. Weyrich, "A procedure for the derivation of project-specific intelligent Digital Twin implementations in industrial automation," May 2022.

[2] B. Ashtari Talkhestani *et al.*, "An architecture of an Intelligent Digital Twin in a Cyber-Physical Production System," *At-Automatisierungstechnik*, Sep. 2019, doi: 10.1515/AUTO-2019-0039.

[3] Y. Xia, M. Shenoy, N. Jazdi, and M. Weyrich, "Towards autonomous system: flexible modular production system enhanced with large language model agents," in *2023 IEEE 28th ETFA*, 2023. doi: 10.1109/ETFA54631.2023.10275362.

[4] Y. Xia, J. Zhang, N. Jazdi, and M. Weyrich, "Incorporating Large Language Models into Production Systems for Enhanced Task Automation and Flexibility," Jul. 2024, doi: 10.48550/arXiv.2407.08550.

[5] Y. Xia, Z. Xiao, N. Jazdi, and M. Weyrich, "Generation of Asset Administration Shell With Large Language Model Agents: Toward Semantic Interoperability in Digital Twins in the Context of Industry 4.0," *IEEE Access*, vol. 12, pp. 84863–84877, 2024, doi: 10.1109/ACCESS.2024.3415470.

[6] P. Häbig *et al.*, "A Modular System Architecture for an Offshore Off-grid Platform for Climate neutral Power-to-X Production in H2Mare," May 2023, [Online]. Available: https://arxiv.org/abs/2305.16285v1

[7] J. Wei, "Chain of Thought Prompting Elicits Reasoning in Large Language Models," *NeurIPS*, 2022.

[8] S. Yao *et al.*, "Tree of Thoughts: Deliberate Problem Solving with Large Language Models," *NeurIPS*, 2023, doi: 10.48550/ARXIV.2305.10601.

[9] S. Yao *et al.*, "ReAct: Synergizing Reasoning and Acting in Language Models," Oct. 2022, [Online]. Available: https://arxiv.org/abs/2210.03629v3

[10] J. Schulman, F. Wolski, P. Dhariwal, A. Radford, and O. K. Openai, "Proximal Policy Optimization Algorithms," Jul. 2017, [Online]. Available: https://arxiv.org/abs/1707.06347v2

[11] R. Rafailov, A. Sharma, E. Mitchell, C. D. Manning, S. Ermon, and C. Finn, "Direct Preference Optimization: Your Language Model is Secretly a Reward Model," *Advances in NeurIPS*, Dec. 2023.

[12] J. C. Sung Park Joseph O *et al.*, "Generative Agents: Interactive Simulacra of Human Behavior," *UIST 2023 - Proc. of the 36th Annual ACM Symp. on User Interface Software and Technology*, Oct. 2023.

[13] Y. Li, Y. Zhang, and L. Sun, "MetaAgents: Simulating Interactions of Human Behaviors for LLM-based Task-oriented Coordination via Collaborative Generative Agents," Oct. 2023, [Online]. Available: https://arxiv.org/abs/2310.06500v1

[14] W. Ma *et al.*, "Large Language Models Play StarCraft II: Benchmarks and A Chain of Summarization Approach," Dec. 2023, [Online]. Available: https://arxiv.org/abs/2312.11865v1

[15] T. Guo *et al.*, "Large Language Model based Multi-Agents: A Survey of Progress and Challenges," Jan. 2024, Accessed: May 24, 2024. [Online]. Available: https://arxiv.org/abs/2402.01680v2